\documentclass[letterpaper,10pt,journal,twoside]{IEEEtran}

\IEEEoverridecommandlockouts

\makeatletter

\let\proof\@undefined
\let\endproof\@undefined
\makeatother

\usepackage[T1]{fontenc}
\usepackage{times}
\usepackage{amsmath,amssymb,amsthm,amsfonts,mathtools}
\usepackage{array}
\usepackage{soul}
\usepackage{tikz}
\usepackage{tikzscale}
\usetikzlibrary{shapes,fit,arrows,positioning,calc}
\usepackage{bm}
\providecommand{\bm}{\pmb}
\usepackage{graphicx}
\usepackage{subfig}
\usepackage{caption}
\usepackage[bookmarks]{hyperref}
\hypersetup{
    colorlinks,
    citecolor=black,
    filecolor=black,
    linkcolor=black,
    urlcolor=black,
    pdfauthor={},
    pdfsubject={},
    pdftitle={A Morphing-Designed Hexarotor Prototype combining Practical Resilience and Efficiency}
}
\usepackage{url}
\usepackage{cite}
\usepackage{paralist}
\usepackage{enumitem}

\graphicspath{{images/}{images/torque_space/}}

\theoremstyle{definition}

\theoremstyle{remark}

\definecolor{myred}{rgb}{0.8500, 0.3250, 0.0980}
\definecolor{myblue}{rgb}{0    0.4470    0.7410}
\definecolor{mygreen}{rgb}{0.4660, 0.6740, 0.1880}
\definecolor{mypurple}{RGB}{139,0,139}

\usepackage[normalem]{ulem}

\newcommand{\vect}[1]{\bm{#1}}		
\newcommand{\matr}[1]{\bm{#1}}		
\newcommand{\nR}[1]{\mathbb{R}^{#1}}		
\newcommand{\SO}[1]{SO({#1})}		
\newcommand{\upperRomannumeral}[1]{\uppercase\expandafter{\romannumeral#1}}	

\newcommand{\frameV}{\mathcal{F}}		
\newcommand{\origin}{O}						
\newcommand{\vX}{\vect{x}}					
\newcommand{\vY}{\vect{y}}					
\newcommand{\vZ}{\vect{z}}					
\newcommand{\vE}[1]{\vect{e}_{#1}}			
\newcommand{\pos}{\vect{p}}				
\newcommand{\dpos}{{\vect{v}}}		
\newcommand{\ddpos}{\dot{\vect{v}}}	
\newcommand{\rotMat}{\matr{R}}			
\newcommand{\angVel}{\vect{\omega}}				
\newcommand{\angAcc}{\dot{\vect{\omega}}}		
\newcommand{\eye}[1]{\matr{I}_{#1}}		

\newcommand{\frameW}{\frameV_W}			
\newcommand{\originW}{\origin_W}		
\newcommand{\xW}{\vX_W}				
\newcommand{\yW}{\vY_W}				
\newcommand{\zW}{\vZ_W}				

\newcommand{\frameP}[1]{\frameV_{p_{#1}}}			
\newcommand{\originP}[1]{\origin_{p_{#1}}}		
\newcommand{\xP}[1]{\vX_{p_{#1}}}				
\newcommand{\yP}[1]{\vY_{p_{#1}}}				
\newcommand{\zP}[1]{\vZ_{p_{#1}}}				

\newcommand{\massR}{m_R}

\newcommand{\inertiaR}{\matr{J}_R}
\newcommand{\frameR}{\frameV_{R}}			
\newcommand{\originR}{O_{R}}					
\newcommand{\xR}{\vX_{R}}								
\newcommand{\yR}{\vY_{R}}								
\newcommand{\zR}{\vZ_{R}}								
\newcommand{\pR}{\pos_{R}}						
\newcommand{\dpR}{\dpos_R} 						
\newcommand{\ddpR}{\ddpos_{R}}					
\newcommand{\angVelR}{\angVel_R}		
\newcommand{\angAccR}{\angAcc_R}		
\newcommand{\rotMatR}{\rotMat_{R}}			
\newcommand{\drotMatR}{\dot{\rotMat}_{R}}			
\newcommand{\forceR}[1]{\vect{f}_{R#1}}		
\newcommand{\torqueR}[1]{\vect{\tau}_{R#1}}		

\newcommand{\thrusti}[1]{\vect{f}_{#1}}
\newcommand{\momenti}[1]{\vect{\tau}_{#1}}
\newcommand{\thrustDir}[1]{\zP{#1}}
\newcommand{\posProp}[1]{\vect{p}_{#1}}
\newcommand{\propVel}[1]{\omega_{#1}}
\newcommand{\cfi}[1]{c_{f_{#1}}}
\newcommand{\cti}[1]{c_{\tau_{#1}}}
\newcommand{\ui}[1]{u_{#1}}
\newcommand{\inputU}{\vect{u}} 
\newcommand{\FMatrix}[2]{{}^{#2} \vect{F}_1^{#1}}
\newcommand{\MMatrix}[2]{{}^{#2} \vect{F}_2^{#1}}

\author{Murat Bronz$^{1 \star}$, Mahmoud Hamandi$^{2}$, Elgiz Baskaya$^1$, Chiara Gabellieri$^{4}$, Antonio Franchi$^{3,4}$%
    \thanks{$^1$Ecole Nationale de l'Aviation Civile, Toulouse, France {\tt \footnotesize \href{mailto:murat.bronz@enac.fr}{murat.bronz@enac.fr}, \href{mailto:elgiz.baskaya@enac.fr}{elgiz.baskaya@enac.fr}}}
    \thanks{$^2$Center for Artificial Intelligence and Robotics, New York University Abu Dhabi, Saadiyat Island, 129188, Abu Dhabi, UAE, {\tt\footnotesize mahmoud.hamandi@nyu.edu}}
    \thanks{$^3$Department of Computer, Control and Management Engineering, Sapienza University of Rome, 00185 Rome, Italy}
    \thanks{$^4$Robotics and Mechatronics lab, Faculty of Electrical Engineering, Mathematics \& Computer Science, University of Twente, Enschede, The Netherlands {\tt \footnotesize \href{mailto:c.gabellieri@utwente.nl}{c.gabellieri@utwente.nl}}, {\tt \footnotesize \href{mailto:schol@r-franchi.eu}{schol@r-franchi.eu}}}
    \thanks{\hspace{-1.0em}This work was partially funded by the European Commission under the project Horizon Europe research agreement no. 101120732 (AUTOASSESS) \& by the ENAC--Airbus--Sopra Steria Drones \& UTM Research Chair.}
}

\title{A Morphing-Designed Hexarotor Prototype combining Practical Resilience and Efficiency}

\begin{document}

\maketitle

\begin{abstract}
This work demonstrates experimentally the existence of a hexarotor prototype, termed \emph{Opti-Hexa}, that simultaneously achieves practical resilience to single-propeller failures and energy efficiency comparable to a standard Star-shaped prototype with the same size, weight, hardware and software.
Leveraging a novel open-source morphing platform, we investigate the trade-offs across a continuous range of geometries by varying the angles between adjacent propellers.
We study practical efficiency through a data-fitted empirical power model and evaluate practical resilience by comparing the position accuracy and rotational kinetic energy during failure to those observed under nominal hovering conditions.
Our experiments confirm the existence of a geometric viability region for this specific morphing platform, where resilience is ensured without the aerodynamic efficiency losses typically associated with practically resilient designs found in the state of the art.
The complete hardware and software of the morphing platform are released to support further research.
\end{abstract}

\begin{IEEEkeywords}
Aerial Systems: Mechanics and Control, Aerial Systems: Applications, Failure Detection and Recovery
\end{IEEEkeywords}

\section*{Supplementary Material}
Open-source design: \href{https://mrtbrnz.github.io/RoBust/}{https://mrtbrnz.github.io/RoBust/}.

\section{Introduction}\label{sec:intro}

Multi-rotor aerial vehicles (MRAVs) are widely used in applications ranging from inspection~\cite{ruggiero2018aerial, 2021g-OllTogSuaLeeFra} to physical interaction~\cite{10558713,10520237} and transportation~\cite{8988166}.
When safety is a priority, MRAVs must be capable of withstanding a single propeller failure. To be considered safe, the vehicle must be able to land with a positioning precision and low rotational kinetic energy comparable to those observed during nominal operation, thereby preventing further damage to the MRAV or its surroundings.
Here, we use the term \emph{resilient} specifically to describe an MRAV possessing these two capabilities, distinct from other possible uses of the term in different contexts.

Among the various multi-rotor designs, quadrotors remain the most common.
However, quadrotors are not resilient; after a single propeller failure, they can at best achieve a dynamic hover characterized by high, and thus unsafe, rotational kinetic energy~\cite{2016-MueDan}.
Designs with five rotors are similarly non-resilient~\cite{2018a-MicRylFra}.
Consequently, six is the minimum number of rotors required to achieve resilience.

However, six rotors are necessary but not sufficient; the geometric arrangement is also critical.
The standard \textit{Star-shaped} hexarotor, where $\gamma = 0$ and $\gamma$ denotes the morphing angle between alternately adjacent arms as illustrated in Fig.~\ref{fig:frame_gamma}, is not resilient~\cite{2018a-MicRylFra}.
It has been proven that resilience can be achieved by shifting the two rotor triangles by a morphing angle $\gamma$, specifically when the configuration satisfies $\gamma \in (0, \pi/6]$~\cite{2018a-MicRylFra}.
This theoretical framework was later generalized in~\cite{2021c-BasHamBroFra}, which demonstrated that resilience is attained if the zero-moment point lies strictly within a design-dependent geometrical set.

\begin{figure}[t]
    \centering
    \includegraphics[width = 0.99\columnwidth]{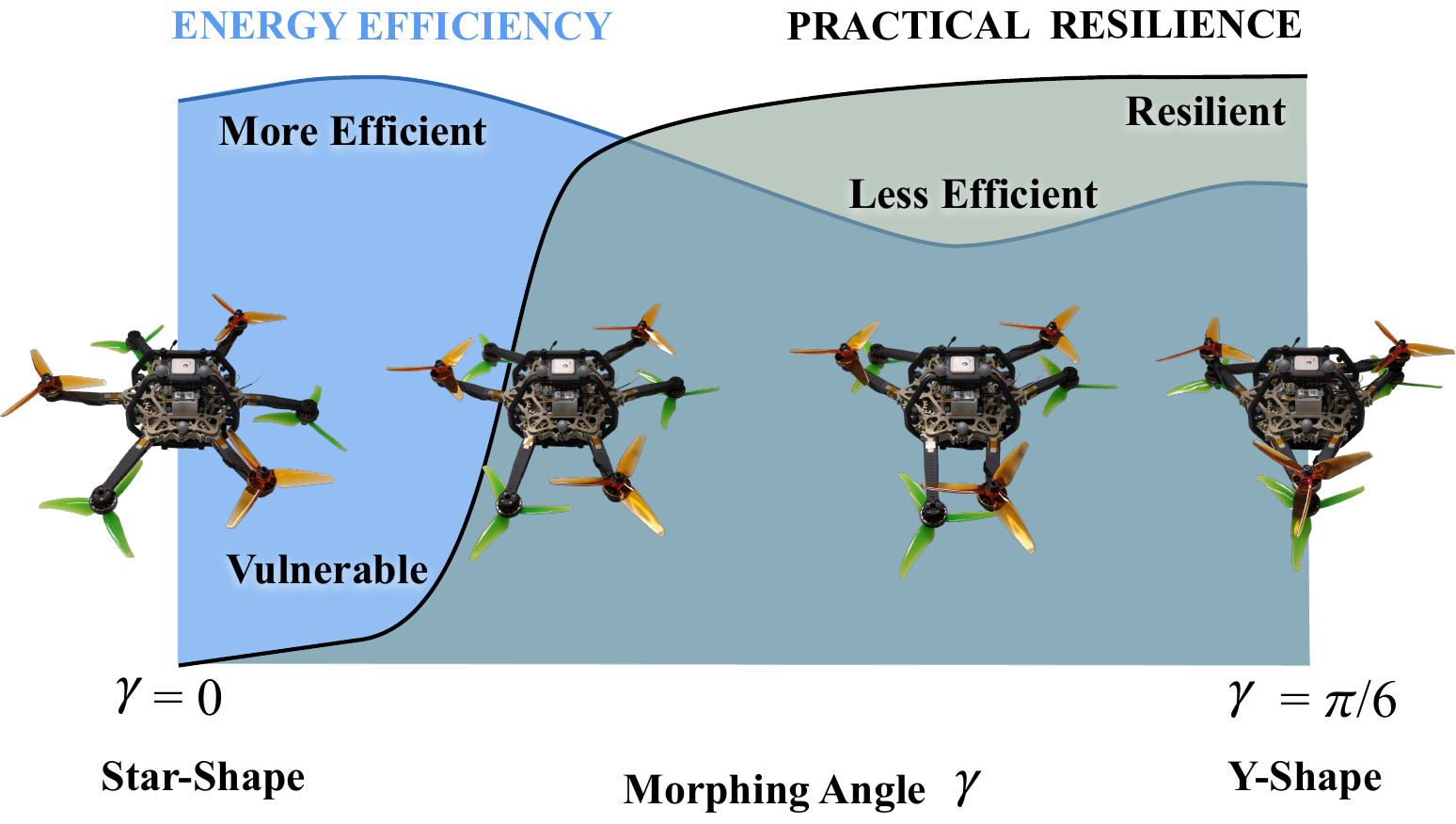}
    \caption{Interplay between energy efficiency and resilience to propeller failures across various configurations, ranging from Star-shaped to Y-shaped hexarotors.}
    \label{fig:mech_design}
\end{figure}

The analyses in~\cite{2018a-MicRylFra,2021c-BasHamBroFra} investigate this property from a purely theoretical perspective. Crucially, they do not account for manufacturing tolerances, unmodeled aerodynamic effects, or other real-world disturbances, necessitating practical experimentation for proper assessment.
We define a platform as \emph{practically resilient} only if its resilience has been experimentally verified.
The first experimental evidence of resilience was presented in~\cite{2021c-BasHamBroFra}, but it was demonstrated solely for the Y-shaped configuration ($\gamma = \pi/6$).
However, the authors also noted that the Y-shape is significantly less efficient than a comparable Star-shaped platform (identical size, weight, and hardware). This inefficiency is due to aerodynamic interference, resulting in a flight duration of only $60\%$ compared to the Star-shaped configuration using the same battery, as shown in~\cite{2021c-BasHamBroFra}.

This work extends this experimental investigation to varying values of $\gamma$, with the goal of identifying and testing the existence of a hexarotor platform that is both practically resilient and \emph{practically efficient}.
We define practical efficiency as the ability to consume hovering power comparable to that of a Star-shaped hexarotor with identical specifications, which serves as our efficiency benchmark.
Consistent with our terminology, the qualifier ``practical'' indicates that this property is validated experimentally.

\subsubsection*{Contribution of this Research}

This work provides, to the best of the authors' knowledge, the first experimental proof of the existence of a hexarotor configuration that is simultaneously practically resilient and practically efficient.
We do not claim that the parameters derived here generalize to all multi-rotor scales or designs; we present an open-source platform to demonstrate that such a physical trade-off is attainable.

To identify this optimal configuration, we employ a \emph{morphing-driven design approach}. We constructed a custom morphing platform to assess the presence of practical resilience and practical efficiency across different values of $\gamma$, ranging between the Star-shaped hexarotor (practically efficient but non-resilient) and Y-shaped hexarotor (practically resilient but non-efficient), as referenced in Fig.~\ref{fig:mech_design}.

Specifically, in this work:
\begin{inparaenum}[i)]
\item \label{contribution_morphing_platform} We introduce an open-source morphing hexarotor platform (hardware and software) capable of continuously transitioning between configurations, allowing us to isolate and study the effects of the morphing angle $\gamma$.
\item \label{contribution_energy} We develop a minimalist 3-parameter model of rotor power consumption as a function of $\gamma$, which is valid specifically for interpolating the data from the developed morphing platform. This analysis reveals an initial plateau of $\gamma$ values where the prototype remains practically efficient.
\item \label{contribution_vulnerability} We analyze the practical resilience of the morphing platform by determining the range of $\gamma$ values for which the translational error and rotational kinetic energy, following a propeller failure, remains comparable to that observed in the absence of failure. This leads to the conclusion that, for this platform, practical resilience is achieved once $\gamma$ exceeds a specific threshold.
\item Ultimately, we synthesize the findings from the previous two points to reveal a key trade-off that establishes the existence of a viability region for this specific morphing design. This yields a hexarotor prototype, named \emph{Opti-Hexa}, which is simultaneously practically resilient and practically efficient.
\end{inparaenum}

The remainder of the paper is structured as follows: Section~\ref{sec:modeling_design} details the prototype modeling and design; Section~\ref{sec:power_consumption} analyzes practical efficiency; Section~\ref{sec:robust} present the study of practical resilience; and Section~\ref{sec:discussion} provides the synthesis, discussion, and conclusions.

\section{The Morphing Prototype}\label{sec:modeling_design}

\begin{figure}[t]
\centering
\includegraphics[width=\columnwidth]{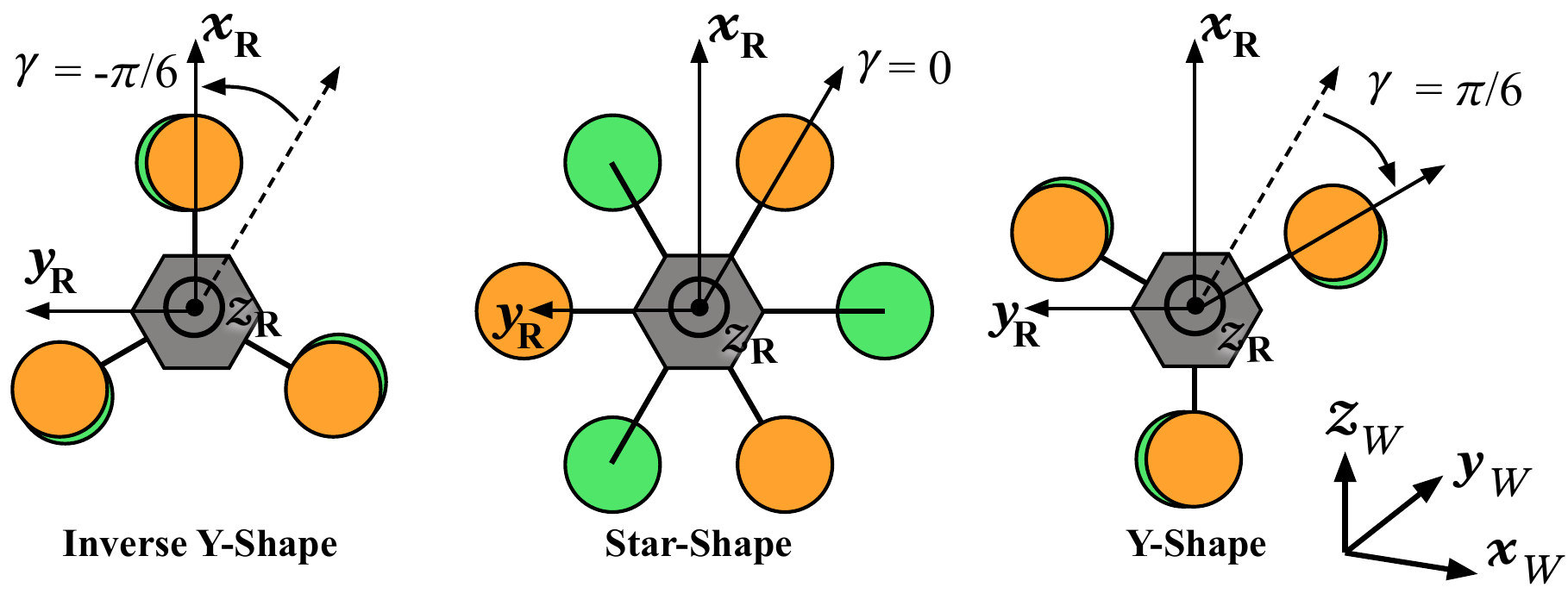}
\caption{Morphing hexarotor prototype configurations with respect to the $\gamma$ angle, with the $\zR$ axis looking upwards.}
\label{fig:frame_gamma}
\end{figure}

Our prototype is a MRAV with six fixed rotors and parallel spinning axes. Let $\frameW = \{\originW ,\xW, \yW, \zW \}$ be the inertial world frame, and let $\frameR = \{\originR, \xR, \yR, \zR \}$ be the body-fixed frame with its origin $\originR$ at the platform's center of mass (CoM), as shown in Fig.~\ref{fig:frame_gamma}.

Let $\pR \in \nR{3}$ and $\dpR \in \nR{3}$ be the position and velocity of $\originR$ in $\frameW$, and $\rotMatR \in \SO{3}$ be the rotation matrix from $\frameW$ to $\frameR$. The angular velocity of $\frameR$ w.r.t. $\frameW$, expressed in $\frameR$, is $\angVel_R$. We have $\drotMatR = \rotMatR [\angVel_R]_\times$, where $[\cdot]_\times$ is the map from $\nR{3}$ to its corresponding skew-symmetric matrix.

Attached to the $i^\text{th}$ motor is the frame $\frameP{i} = \{\originP{i}, \xP{i}, \yP{i}, \zP{i}\}$, where $\originP{i}$ is at the rotor's center, $i=1,\ldots,6$. The position of $\originP{i}$ in $\frameR$ is $\posProp{i} \in \nR{3}$.
The $i^\text{th}$ rotor rotates about $\thrustDir{i}$ with a spinning rate $\propVel{i}\in \nR{}$, where $\propVel{i} > 0$ corresponds to spinning in the direction of $\thrustDir{i}$. To model both rotor types, we set $k_i=1$ for descending-chord rotors and $k_i=-1$ otherwise, where $k_i$ is a binary variable referring to the direction of rotation. The rotor generates a thrust force $\thrusti{i} \in \nR{3}$ and a drag moment $\momenti{i}^d \in \nR{3}$ at $\originP{i}$:
\begin{align}
\thrusti{i} &= k_i \cfi{i} |{\propVel{i}} |\propVel{i} \thrustDir{i},\label{eq:nominal_thrust}\\
\momenti{i}^d &= -\cti{i} |{\propVel{i}}|\propVel{i} \thrustDir{i},
\label{eq:nominal_torque}
\end{align}
where $\cfi{i},\cti{i} \in \nR{}_{>0}$ are the thrust and moment coefficients. The control input is $\ui{i} = |{\propVel{i}}|\propVel{i}$.

The total force $\forceR{}$ and moment $\torqueR{}$ at $\originR$ are:
\begin{align}
\forceR{} &= \sum_{i=1}^{n} k_i\cfi{i} \ui{i} \thrustDir{i},\label{eq:total_force}\\
\torqueR{} &= \sum_{i=1}^{n} (\momenti{i}^f + \momenti{i}^d) = \sum_{i=1}^{n} (k_i\cfi{i} \posProp{i} \times \thrustDir{i} - {\cti{i} \thrustDir{i}}) \ui{i},
\label{eq:total_moment}
\end{align}
where $\momenti{i}^f$ denotes the \textit{thrust-induced moment} generated by the thrust force of the $i$-th rotor acting about the vehicle center of mass. As customary, we assume identical coefficients for all rotors, with odd-indexed rotors being descending-chord ($k_i=1$) and even-indexed ascending-chord ($k_i=-1$).

The prototype is parametrized by the morphing angle ${\gamma\in[- \frac{\pi}{6},\frac{\pi}{6}]}$. This angle is the uniform rotation applied to each arm around $\zR$, relative to the neutral Star-shaped configuration. Specifically, $\gamma=0$ is the Star-shaped configuration and $\gamma=\pm \frac{\pi}{6}$ are the two antipodal Y-shaped ones, Y-shape $\gamma= \frac{\pi}{6}$ and inverse Y-shape $\gamma= -\frac{\pi}{6}$, as illustrated in Fig.~\ref{fig:frame_gamma}. We include negative $\gamma$ values in our analysis to counterbalance potential biases due to manufacturing tolerances.

Rotors are in two parallel planes to allow for overlap. Arms with even indices $i$ are on the CoM plane, while arms with odd indices $i$ are on a lower plane, displaced by $-l_z$ along $\zR$. This allows partial or full rotor overlap for certain $\gamma$ values.
We have then
\begin{align}
\posProp{i} = \underbrace{\rotMat_z  \left((i-1)\tfrac{2 \pi}{6} - (-1)^i \gamma \right)}_{\rotMat_\gamma(i)}
\left[\begin{smallmatrix}
l_x\\
0\\
-l_z {(i\operatorname{mod}2)}
\end{smallmatrix}\right], 
    \label{eq:positions_props}
\end{align}
where $i=1,\ldots,6$, $\rotMat_z$ is the canonical rotation matrix about $\zR$, $l_x$ is the arm length, $l_z$ is the vertical displacement, and $i\operatorname{mod}2$ is the remainder of $i$ divided by $2$.

Finally, the dynamics of the $\gamma$-parametrized hexarotor are given by the Newton-Euler equations:
\begin{align}
 \begin{split}
 &\left[\begin{smallmatrix}
 \massR\eye{3} & \boldsymbol{0}_{3}^{} \\
 \boldsymbol{0}_{3}^{} & \inertiaR
 \end{smallmatrix}\right]
 \begin{bmatrix}
 \ddpR \\
 \angAccR
 \end{bmatrix}
 = 
 \left[\begin{smallmatrix}
 -\massR g \vect{e}_{3}^{} \\
 -\angVelR \times   \inertiaR \angVelR 
 \end{smallmatrix}\right] +
 \left[\begin{smallmatrix}
 \rotMatR & \boldsymbol{0}_{3}^{} \\
 \boldsymbol{0}_{3}^{} & \eye{3}
 \end{smallmatrix}\right]
 \begin{bmatrix}
 \forceR{}  \\
 \torqueR{} 
 \end{bmatrix}= \\
 & = \begin{bmatrix}
 -\massR g \vect{e}_{3}^{} \\
 -\angVelR \times   \inertiaR \angVelR 
 \end{bmatrix} +
 \begin{bmatrix}
 \rotMatR & \boldsymbol{0}_{3}^{} \\
 \boldsymbol{0}_{3}^{} & \eye{3}
 \end{bmatrix}
 \matr{F}(\gamma)\inputU,
 \end{split}
 \label{eq:dyn}
 \end{align}
where $\massR$ and $\inertiaR$ are the vehicle's mass and inertia matrix; $g$ is gravity, $\vE{i}$ is the $i^\text{th}$ column of $\eye{3};$ $\inputU=[\ui{1} \cdots \ui{6}]^\top,$ and $\matr{F}(\gamma)$ is the allocation matrix from substituting \eqref{eq:total_force} and \eqref{eq:total_moment} into~\eqref{eq:dyn}. We decompose $\matr{F}(\gamma)$ as $\matr{F}(\gamma)=\begin{bmatrix}
\FMatrix{}{}\\\MMatrix{}{}(\gamma)
\end{bmatrix}$, where
\begin{align}
\FMatrix{}{} = \cfi{}
\begin{bmatrix}
0 & 0 & 0 & 0 & 0 & 0 \\
0 & 0 & 0 & 0 & 0 & 0 \\
1 & 1 & 1 & 1 & 1 & 1 \\
\end{bmatrix},
\end{align}
\begin{align}
\small
\MMatrix{}{}(\gamma) =
\bgroup
\!\begin{aligned}
\cti{}&\left[
\begin{smallmatrix}
rs(\frac{\pi}{6}-\frac{\gamma}{2}) & +rs(\frac{\pi}{2} + \frac{\gamma}{2}) & +rs(\frac{5\pi}{6}-\frac{\gamma}{2}) \\ 
-rc(\frac{\pi}{6}-\frac{\gamma}{2}) & -rc(\frac{\pi}{2} + \frac{\gamma}{2}) & -rc(\frac{5\pi}{6} - \frac{\gamma}{2}) \\
-1 & 1 & -1 \\
\end{smallmatrix}\right.\\
&\qquad
\left.\begin{smallmatrix}
rs(\frac{7\pi}{6}+\frac{\gamma}{2}) & +rs(\frac{3\pi}{2} - \frac{\gamma}{2}) & +rs(\frac{11\pi}{6}+\frac{\gamma}{2})  \\
- rc(\frac{7\pi}{6} +\frac{\gamma}{2}) & -rc(\frac{3\pi}{2} - \frac{\gamma}{2}) & -rc(\frac{11\pi}{6} +\frac{\gamma}{2}) \\
1 & -1 & 1 \\
\end{smallmatrix}
\right],
\end{aligned}
\egroup
\label{eq:F_2}
\end{align}
where $r = (\cfi{} / \cti{} ) l$ with $l$ the vehicle arm length, $s(\cdot) = \sin(\cdot)$, and $c(\cdot) = \cos(\cdot)$.

\begin{figure}[t]
    \centering
    \includegraphics[width = 0.99\columnwidth]{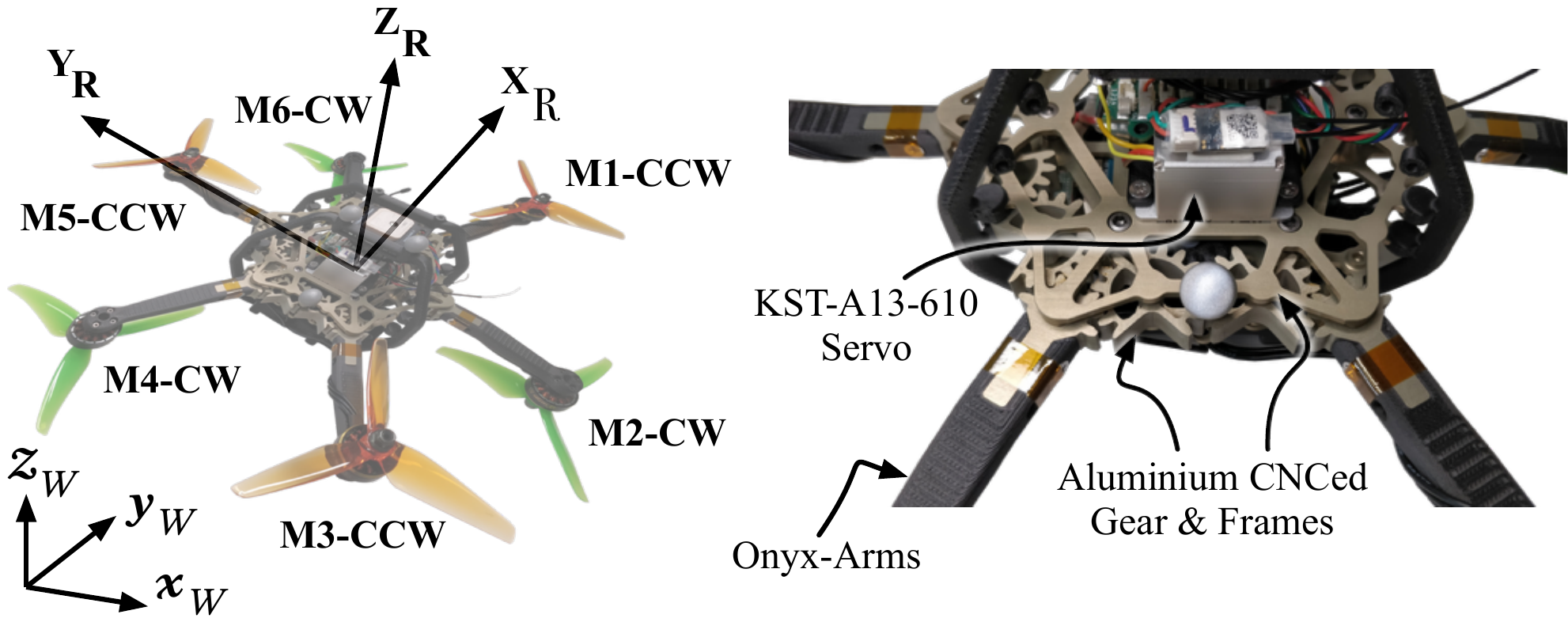}
    \caption{The open-source morphing hexarotor design is shown. Left: The manufactured platform. Right: Exploded view of the design showing the gear mechanism. The platform is able to morph between Star- and Y-shaped configurations. The design files are available as open-source in the supplementary material.}
    \label{fig:frames_and_design}
    \vspace{-0.4cm}
\end{figure}

The built prototype is shown in Fig.~\ref{fig:frames_and_design}.
The morphing mechanism comprises two CNC-milled aluminum plates for the upper and lower body, and six articulated arms with 3D-printed sections using Onyx\footnote{\url{https://markforged.com/materials/plastics/onyx}} material, connected to aluminum gears with horizontal screws, as shown in Fig.~\ref{fig:frames_and_design}.
Fourteen articulation gears, fabricated using wire electrical discharge machining, rotate consecutive arms in opposite directions synchronously. All aluminum parts are anodized for corrosion resistance.
The configuration change is actuated by two KST-A13-610 servos, which are internally modified to obtain potentiometer readings for absolute angular position information, thus measuring the current value of $\gamma$.
The corresponding arm position information is provided to the on-board autopilot to determine the geometric configuration, thereby continuously updating the matrix $\mathbf{F}(\gamma)$ used by the on-board attitude controller at all times.

The propulsion system consists of T-motor F2203.5 (Kv 1500) motors paired with T-5150 propellers, powered by a 14.6\,V (four-cell) battery and T-motor Velox V2 speed controllers.
This setup provides a total thrust-to-weight ratio of approximately $2.5$.
When fully assembled, the platform has a diameter of approximately 0.43\,m and a mass of approximately 750\,g, including the 190\,g battery.

The open-source \emph{Paparazzi Autopilot} \cite{hattenberger2014using} is used both for hardware (Tawaki-V1.0) and software.
The on-board attitude control and guidance methods rely on Incremental Non-linear Dynamic Inversion (INDI)~\cite{smeur2018cascaded}, which is a model-free controller, requiring only the actuator models, suitable for the challenging morphing capabilities of the prototype. While the controller is unchanged throughout the experiments, INDI estimates online the efficiency matrix, and allocates thrust inputs accordingly.

The platform is available at: \href{https://mrtbrnz.github.io/RoBust/}{https://mrtbrnz.github.io/RoBust/}.

\section{Region of Practical Efficiency}\label{sec:power_consumption}

This section details the methodology used to measure and interpolate the thrust and power consumption as a function of the morphing angle $\gamma$ in order to assess the region in which the morphing prototype is practically efficient.

\subsection{Proposed Primary $\gamma$-dependent Effects on the Thrust}
\label{sec:power_cons_model}

Aerodynamic interaction can reduce the total thrust of overlapping or blocked propellers, affecting the energy consumed by the UAV. The thrust behavior is characterized by fitting the experimental test data to an analytical model. We identified three primary $\gamma$-dependent effects based on the literature from~\cite{2015-Manikandan,2021-Stokkermans}, in addition to our experimental observations.

\subsubsection{Motor-and-arm Blockage ($c_{maa}$)}
The portion of the arms and motors beneath the upper propeller disc varies with $\gamma$, which can affect the propeller slipstream. This effect evolves as the lower motor and arm enter and then overlap with the upper components, resulting in changes to the total blocked area.

\subsubsection{Propeller Overlap ($c_{prop}$)}
The $\gamma$-dependent geometric intersection area between the lower and upper propeller discs is expected to influence both propellers. The lower propeller experiences a reduced local angle of attack due to the upper propeller's inflow, while the upper slipstream is partially obstructed by the lower propeller.

\subsubsection{Tip-to-Tip Interaction ($c_{t2t}$)}
The $\gamma$-dependent proximity of propeller tips is expected to influence thrust. Close proximity can increase thrust via tip vortex interaction, which is consistent with the small efficiency gains observed at certain angles.

\begin{figure}[h]
\centering
\includegraphics[width=\columnwidth]{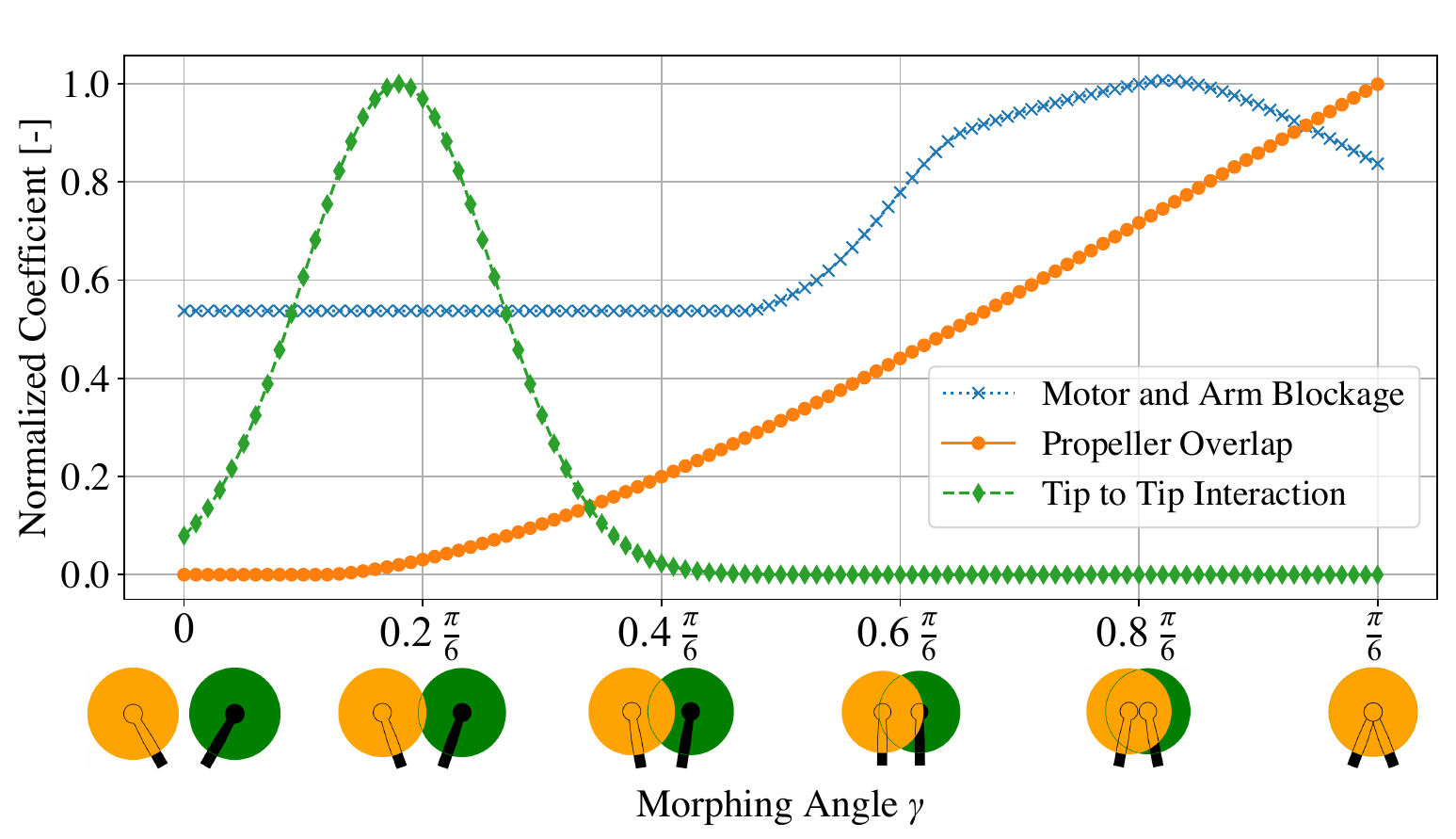}
\caption{Proposed primary $\gamma$-dependent effects on the generated thrust: Motor-and-arm Blockage, Propeller Overlap, and Tip-to-Tip Interaction. Each curve represents the described normalized overlapping area of the corresponding effect. Therefore, the ratio between curves does not indicate any dominance.}
\vspace{-0.7cm}
\label{fig:thrust_primitives}
\end{figure}

Figure~\ref{fig:thrust_primitives} illustrates the expected thrust variation from these three effects as a function of ${\gamma} \in [0, \frac{\pi}{6}]$, with symmetric behavior expected for ${\gamma} \in [-\frac{\pi}{6}, 0]$. The curves represent normalized areas or, for tip-to-tip, a Gaussian function. We denote them $c_{\text{maa}}(\gamma)$, $c_{\text{prop}}(\gamma)$, and $c_{\text{t2t}}(\gamma)$, which are computed from platform geometry. A value of zero means no effect. Since all curves are normalized, their ratios do not indicate dominance. Propeller overlap and blockage reduce thrust, while tip-to-tip interaction increases thrust.

We modify~\eqref{eq:nominal_thrust} to hypothesize that the thrust magnitude from a propeller pair (upper~$(\cdot)_{\text{up}}$ and bottom~$(\cdot)_{\text{bot}}$) is:
\begin{align}
\|\thrusti{\text{up}}+\thrusti{\text{bot}}\| &= c_{f_{\text{up}}}(\gamma) \, \propVel{\text{up}}^2 + c_{f_{\text{bot}}}(\gamma) \, \propVel{\text{bot}}^2,\label{eq:thrust_pair}
\end{align}
with thrust coefficients $c_{f_{\text{up/bot}}}(\gamma)$ modeled as:
\begin{align}
c_{f_{\text{up}}}(\gamma) & \approx c_{\text{nom}} + \lambda_1 c_{\text{maa}}(\gamma) + \lambda_2 c_{\text{prop}}(\gamma) + \lambda_3 \, c_{\text{t2t}}(\gamma), \label{eq:hypothesis1}\\
c_{f_{\text{bot}}}(\gamma) & \approx c_{\text{nom}} + \lambda_2 c_{\text{prop}}(\gamma) + \lambda_3 \, c_{\text{t2t}}(\gamma),
\label{eq:hypothesis2}
\end{align}
where $c_{\text{nom}}$ is the nominal thrust coefficient, $\lambda_1, \lambda_2\in\mathbb{R}_{<0}$, and $ \lambda_3 \in \mathbb{R}_{>0}$ are weighting factors. The following subsections present data supporting this hypothesis.

\begin{figure}[t]
    \centering
    \includegraphics[width = 0.99\columnwidth]{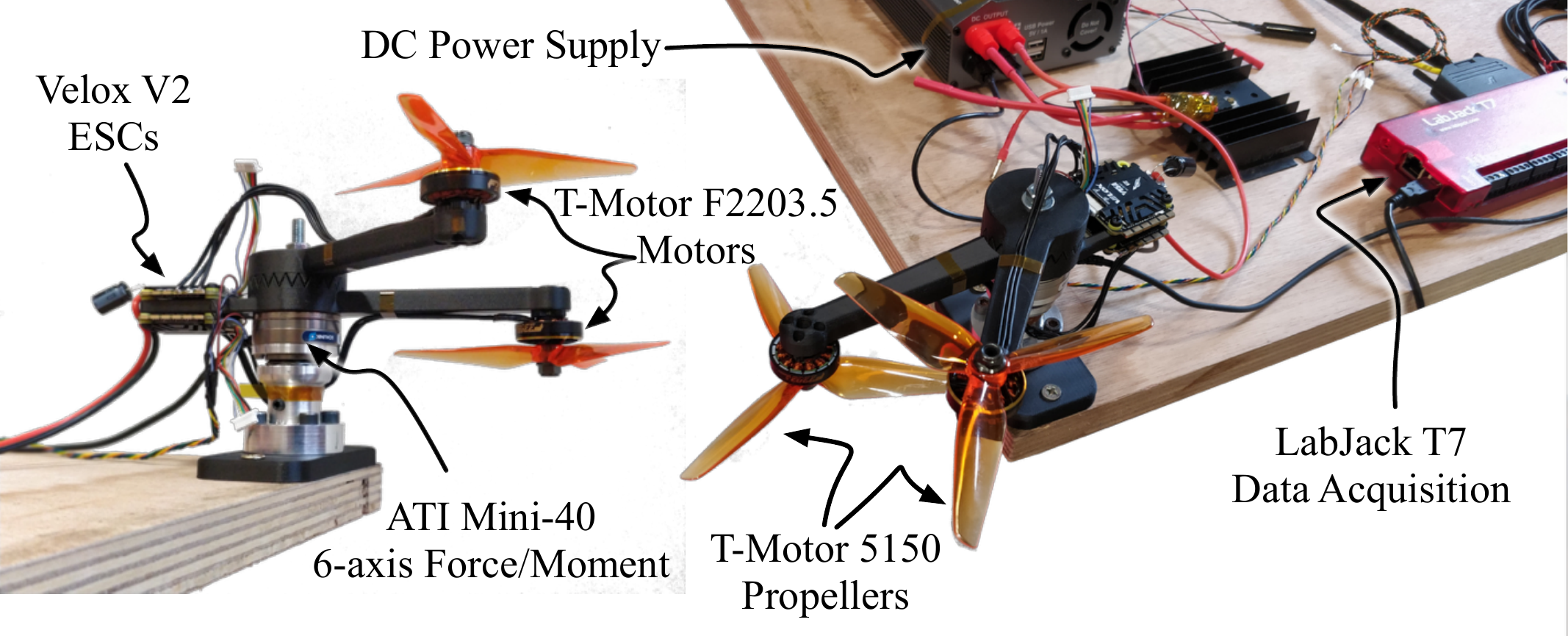}
    \caption{Experimental setup built and used for the investigation of the phenomena involved in the interaction between propellers. The setup is comprised of a force and moment balance, a data acquisition card, electronic speed controllers, and a test bench manually positioning the rotor arm angle for the parametric study.}
    \label{fig:propulsion_test_bench}
\end{figure}

\subsection{Analyses of Force-torque Data Acquired on a Test-bench}

To validate the three effects from Section~\ref{sec:power_cons_model}, we built a custom test bench (Fig.~\ref{fig:propulsion_test_bench}) with an ATI Mini-40 force-torque sensor to test various overlap configurations.

\begin{figure}[t]
    \centering
\includegraphics[width = 0.95\columnwidth]{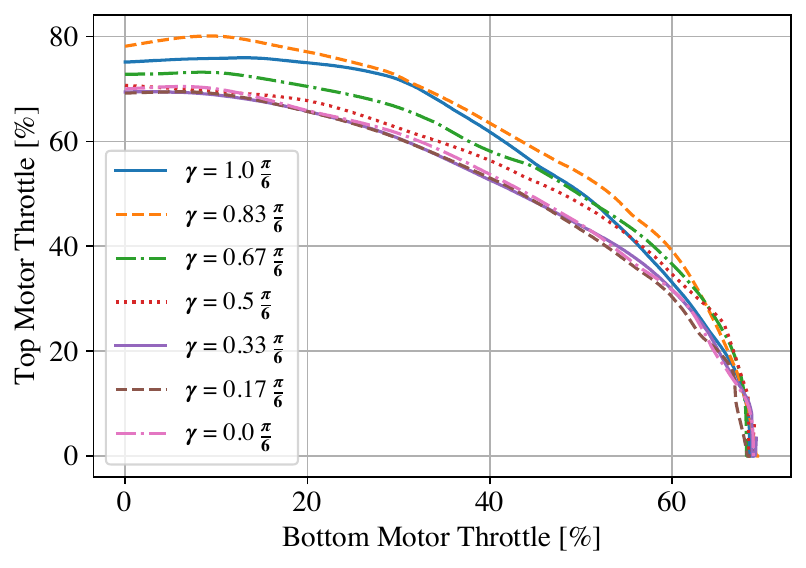}
\caption{Level curves of the total thrust produced by top and bottom propellers at various $\gamma$ values. Each curve represents the throttle combinations where the sum of thrusts reaches $2.67$\,N, corresponding to the required hovering thrust for the pair of propellers.}
    \label{fig:total_thrust}
    \vspace{-0.5cm}
\end{figure}

We focused on the hover condition, requiring $T^{\text{hover}}/3 = 2.45$\,N per rotor pair. Figure~\ref{fig:total_thrust} shows level curves of throttle combinations that produce this thrust, parameterized by $\gamma$.
The curve extremities in Fig.~\ref{fig:total_thrust} (single-rotor operation) validate the motor-and-arm blockage. At the lower-right (top prop off), the bottom propeller's throttle is unaffected by $\gamma$, as hypothesized. Conversely, at the upper-left (bottom prop off), the top propeller's required throttle varies (69\%--79\%), confirming its thrust is influenced by the $\gamma$-dependent blockage shown in Fig.~\ref{fig:thrust_primitives}.
The propeller overlap effect is also confirmed. Near $\gamma = \frac{\pi}{6}$, as the bottom motor's throttle increases, the top's remains unchanged, supporting the hypothesis that overlap maximizes efficiency loss for the lower propeller.

\begin{figure}[t]
    \centering
\includegraphics[width = 0.99\columnwidth]{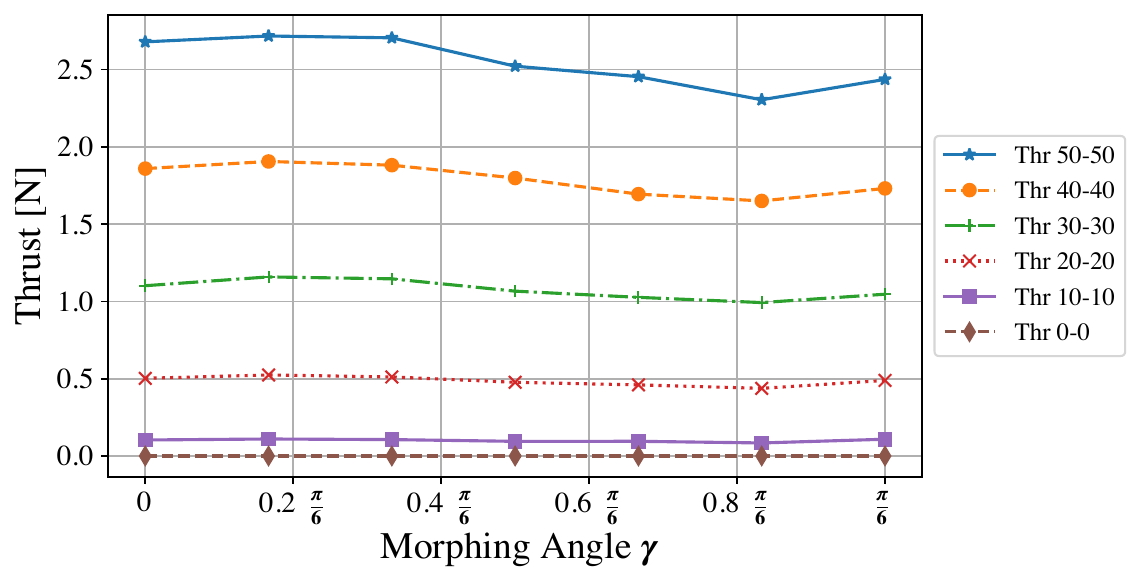}
    \caption{Combined total thrust force produced by the top and bottom propellers for various throttle values is shown as a function of the morphing angle~$\gamma$.}
    \label{fig:thrust}
    \vspace{-0.5cm}
\end{figure}

Finally, Fig.~\ref{fig:thrust} plots total thrust for a propeller pair at equal throttles vs. $\gamma$. The maximum thrust occurs when the propeller tips are closest, near $|{\gamma}| = 0.2\frac{\pi}{6}$, supporting the tip-to-tip interaction effect. This analysis predicts that minimum energy consumption is not at the Star-shaped configuration ($\gamma = 0$), a finding corroborated by independent flight data in the following section (Section~\ref{sec:flight_data}).

\subsection{Analyses of Power-consumption Data Acquired in Flight}
\label{sec:flight_data}

\begin{figure}[t]
    \centering
\includegraphics[width = 0.99\columnwidth]{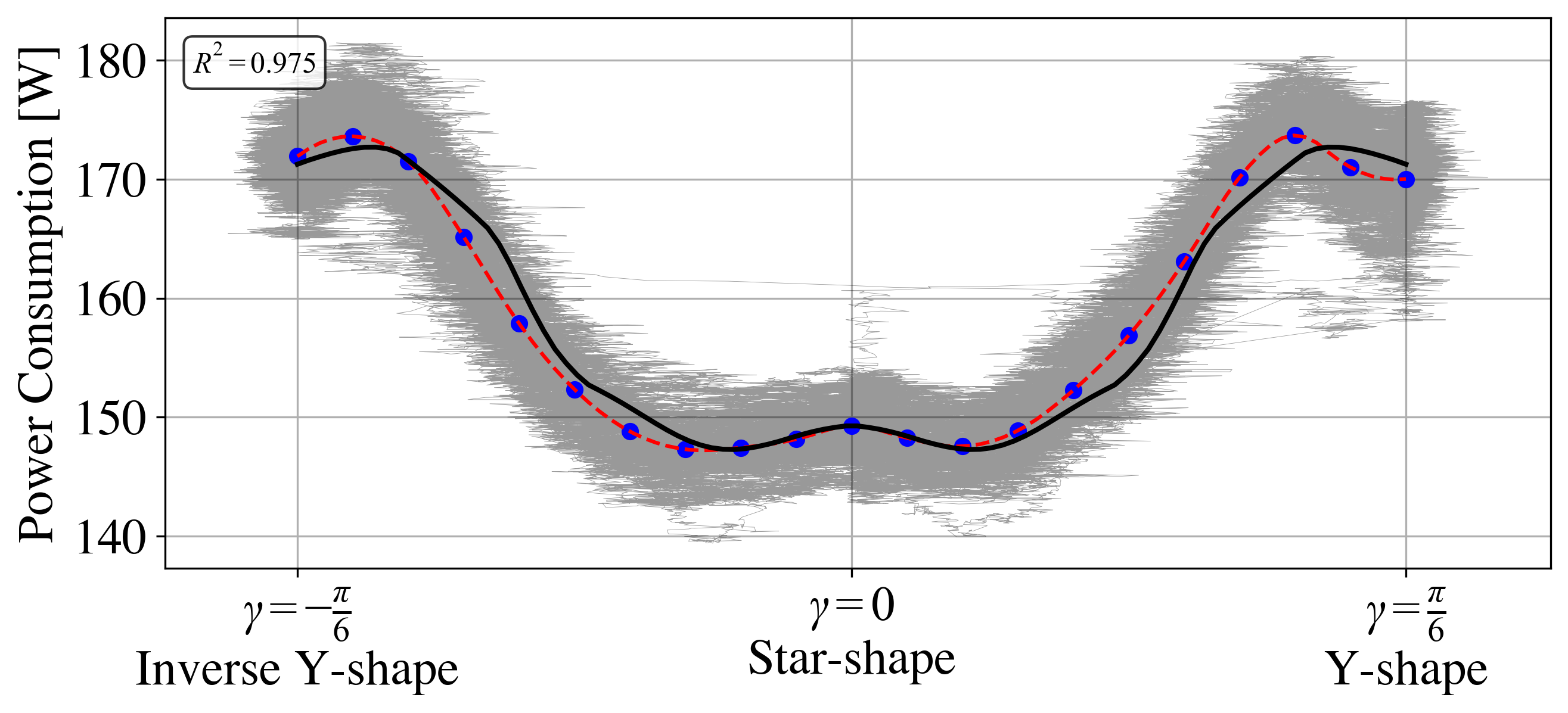}
    \caption{Power consumption of the whole platform for different values of $\gamma$: in gray the scattered raw data of 27 minutes of flight (3 separate flights recorded at 100 Hz); in blue round marks averaged data points, in dashed red line a spline fit of the averages, and in solid black line the fitting of the 3-parameter power model from physical primitives are shown.}
    \label{fig:power_consumption}
\end{figure}

We acquired 27~minutes of hovering flight data at the VTO\footnote{\url{https://www.enac.fr/en/drone-flight-arena-toulouse-occitanie-0}} flight arena, where the drone was commanded to a desired fixed position and orientation.
Position and orientation were captured by a motion capture system. During three consecutive flights, ${\gamma}$ was autonomously varied at an angular rate of 0.06 rad/s.
The raw power data is shown in gray in Fig.~\ref{fig:power_consumption}, along with averaged data (blue) and an interpolating spline (red).
To validate the 3-parameter thrust model from \eqref{eq:hypothesis1} and \eqref{eq:hypothesis2}, we must relate the coefficients $c_{f_{\text{up}}}(\gamma)$ and $c_{f_{\text{bot}}}(\gamma)$ to the measured power consumption. This relationship allows identifying $\lambda_1$, $\lambda_2$, and $\lambda_3$ via regression.

Our robust flight controller compensates for thrust variations by adjusting motor throttle to maintain the desired thrust. Consequently, power consumption is an indirect function of $\gamma$. We now derive this relationship to estimate the $\lambda$ parameters.

Due to the slow $\gamma$ variation, we assume constant propeller rotation rates in short intervals. With negligible motor friction, power $p_i$ counteracts drag, following~\eqref{eq:nominal_torque}. Power is the product of torque and rotation rate:
\begin{align}
p_i = \cti{} |\propVel{i}|^3.
\label{eq:power_drag}
\end{align}
Hence, for a pair of motors, the total power consumption is
\begin{align}
p_{\text{total}} = p_{\text{up}} + p_{\text{bot}} = \cti{} |\propVel{\text{up}}|^3 + \cti{}|\propVel{\text{bot}}|^3.
\label{eq:pair_power_drag}
\end{align}
The propeller rotation rates $\omega_{\text{up}}$ and $\omega_{\text{bot}}$ are determined by a control allocation mapping $\mathcal{K}$.
Since thrust estimation depends on $c_{f}(\gamma)$, a function of $\lambda_1, \lambda_2, \lambda_3$, the rates $\omega_{\text{up}}$ and $\omega_{\text{bot}}$, and thus $p_{\text{total}}$, are also functions of $\gamma$ and the $\lambda$ parameters.

Assuming symmetric thrust generation ($ \|\thrusti{\text{up}}+\thrusti{\text{bot}}\| = \massR g/3$), we summarize the controller's actions as:
\begin{equation}
\mathcal{K}(\gamma,\lambda_1, \lambda_2, \lambda_3)|_{\text{thrust} = \massR g/3} = (\omega_{\text{up}} , \omega_{\text{bot}})|_{\text{thrust} := \massR g/3},
\label{eq:omega_from_controller}
\end{equation}
which does not constitute a closed-form solution.
However, \eqref{eq:omega_from_controller} can be run as a black box (offline controller simulation) to find the total estimated power $p_{\text{total}}(\lambda_1, \lambda_2, \lambda_3, \gamma)$ by substituting the resulting rates into~\eqref{eq:pair_power_drag}.

We use this to set up a standard data-fitting problem to find the coefficients $\lambda$ that minimize the error between measured and modeled power:
\begin{equation}
\min_{\lambda_1, \lambda_2, \lambda_3} \sum_{\gamma=-1}^{1} \left( p_{\text{measured}}(\gamma) - p_{\text{total}}(\lambda_1, \lambda_2, \lambda_3, \gamma) \right)^2,
\end{equation}
subject to:
\begin{align}
&(\omega_{\text{up}} , \omega_{\text{bot}}) = \mathcal{K}(\gamma,\lambda_1, \lambda_2, \lambda_3),\\
&\|\bm{f}_{\text{up}} + \bm{f}_{\text{bot}}\| = \massR g/3, \quad \forall \gamma \in [-\pi/6,\pi/6],\\
&\|\bm{f}_{\text{up}}\| = c_{f_{\text{up}}}(\gamma) \, \omega_{\text{up}}^2, \quad \|\bm{f}_{\text{bot}}\| = c_{f_{\text{bot}}}(\gamma) \, \omega_{\text{bot}}^2.
\end{align}
This problem is solved numerically using a standard Nelder-Mead search method to find the best-fitting $\lambda$ parameters ($\lambda_1=-0.54$, $\lambda_2=-0.05$, $\lambda_3=0.012$).

The resulting data-fitting is the black curve in Fig.~\ref{fig:power_consumption}. Despite having only three parameters, the model accurately captures the main features of the measured power data. Namely, the black curve has peaks for values of $\gamma$ far from zero, with a slight decrease towards the extreme values of $\gamma=\pm\frac{\pi}{6}$, and with decreasing values when $\gamma$ approaches zero and a local maximum around $\gamma=0$.

Consistent with the experimental findings reported in~\cite{2021c-BasHamBroFra}, Fig.~\ref{fig:power_consumption} confirms the Star-shape ($\gamma = 0$) is more efficient than the Y-shape ($\gamma = \pm \frac{\pi}{6}$). While the overall trend follows the anticipated monotonic decrease in power consumption from the Y-shape to the Star-shape, the data exhibits two local deviations from this behavior. The first suggests that the most efficient configuration is slightly away from $\gamma = \pm \frac{\pi}{6}$ instead of being exactly at $\gamma = \pm \frac{\pi}{6}$, and can be explained by motor-and-arm blockage. The second, an initial power reduction near $\gamma = 0$, is consistent with the tip-to-tip interaction effect.
Finally, the distribution is not perfectly symmetrical, justifying our design choice to test both positive and negative $\gamma$ values and use them all during the fitting process for an unbiased analysis.

\subsection{Key Insights and Main Result}

We hypothesized that three primary effects are sufficient to fit the power consumption of our prototype. Experimental evidence from two independent sources (test bench and in-flight power) strongly supports this hypothesis. While our simplified model does not perfectly match the spline fit in Fig.~\ref{fig:power_consumption}, it captures all key features. We found the model to be: 1) explainable: it is grounded in physical principles, providing insight into the observed data; 2) parsimonious: it uses only three parameters, making it less susceptible to overfitting; and 3) adaptable: the physics-based framework provides a methodology that could, in theory, be extended to analyze other UAV geometries in future work.

\subsubsection*{Main Result}
Based on the experimental data points and the interpolation provided by the proposed model, we conclude that the morphing prototype remains practically efficient for any value of $\gamma$ within the interval $(-\pi/18, \pi/18)$, corresponding to $(-10^\circ, 10^\circ)$.
The next step is to verify whether, within this interval, there exist configurations that are also practically resilient.

\section{Region of Practical Resilience}\label{sec:robust}

This section details the design and results of the experimental campaign conducted to identify the range of $\gamma$ values for which the morphing prototype exhibits practical resilience.
As defined in Section~\ref{sec:intro}, a physical platform is defined as practically resilient only if experimental evidence demonstrates that its rotational kinetic energy and positioning accuracy, following the loss of a single propeller at hover, remain comparable to those observed under nominal operating conditions.
Given that the platform is underactuated, measuring the full rotational kinetic energy is unnecessary; high positioning accuracy during hover inherently implies a consistent rotational kinetic energy about the horizontal body axes.
Therefore, to assess practical resilience, it suffices to monitor the following two functions: 1) the translational error, $E^{\text{tran}}$, and 2) the rotational kinetic energy about the vertical body axis, $E_{\psi}^{\text{rot}}$.
These are defined as:
\begin{align}
    E^{\text{tran}} &= \frac{1}{2} m (\mathbf{e}_{\mathbf{p}}^\top \mathbf{e}_{\mathbf{p}} + \dot{\mathbf{e}}_{\mathbf{p}}^\top \dot{\mathbf{e}}_{\mathbf{p}}),\\
    E_{\psi}^{\text{rot}} &= \frac{1}{2} I_z \omega_{\psi}^{2},
\end{align}
where $\mathbf{e}_{\mathbf{p}} = \mathbf{p}_d - \mathbf{p}$ represents the positional error, $\omega_{\psi}$ denotes the yaw rate, and $I_z$ is the moment of inertia about the $z$-axis.
To ensure safety during testing, a supervisory layer was implemented to force a rapid transition to the stable Y-shape configuration if energy thresholds were exceeded.

Three types of experiments were conducted.

\subsubsection{Type 1: Continuous Sweep under Failure}
The hexarotor is initialized in the Y-shape configuration ($\gamma = \pi/6$). A single propeller is intentionally stopped while hovering. The platform then gradually transitions $\gamma$ to the Inverse Y-shape ($\gamma = -\pi/6$) over a duration of 30 seconds, passing through the Star-shape ($\gamma = 0$). This procedure is repeated for all six propellers.

\subsubsection{Type 2: Discrete Static Failure Tests}
The hexarotor hovers at a fixed configuration $\gamma$. A propeller failure is induced and maintained for 10 seconds. This test is repeated for every propeller and for each configuration in the set $\Gamma$, defined as:
\[
\Gamma = \left\{0, \pm 0.1, \pm 0.2, \pm 0.3, \pm 0.6, \pm 0.8, \pm 1\right\} \times \frac{\pi}{6} \, [\text{rad}].
\]

\subsubsection{Type 3: Dynamic Flight under Failure}
This experiment extends the analysis to dynamic trajectories. The hexarotor tracks a circular path while its configuration cycles through $\gamma \in \Gamma$. For each configuration, a 10-second propeller failure is induced, cycling through all propellers.

\begin{figure}[t]
    \centering
    \includegraphics[width = 1.0\columnwidth]{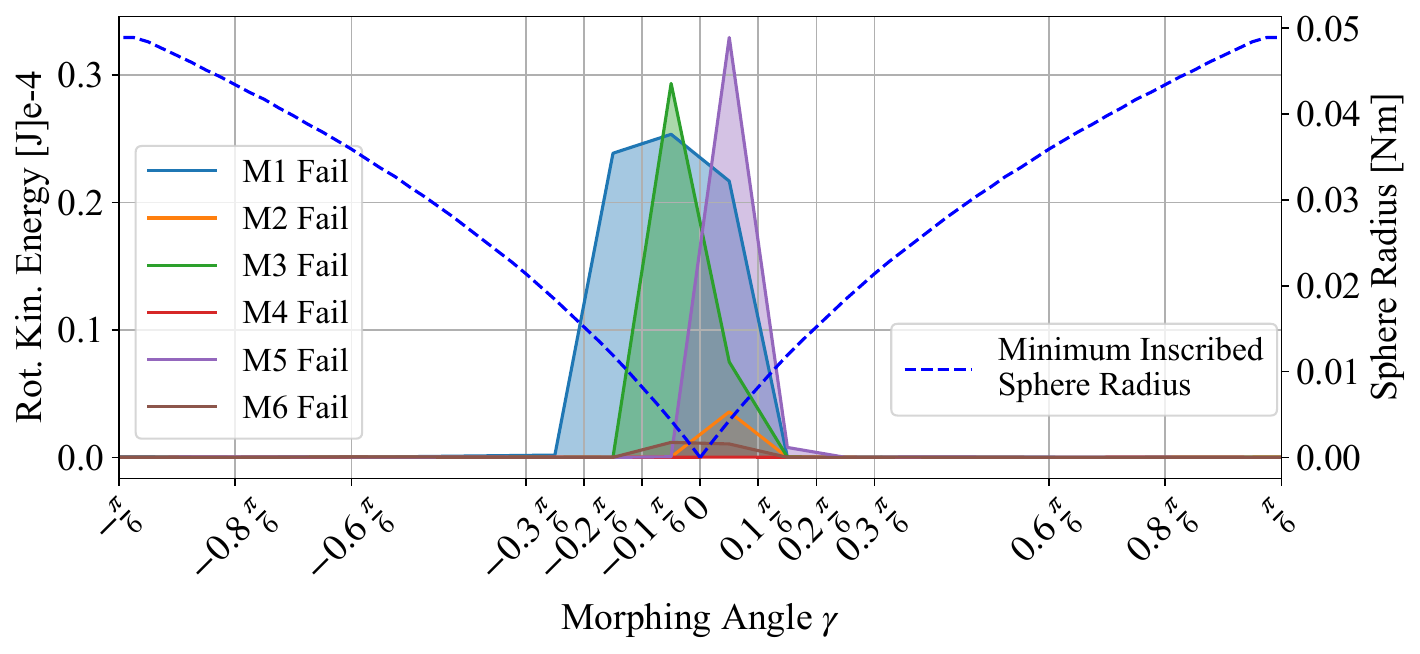}
    \caption{Average rotational kinetic energy $E_{\psi}^{\text{rot}}$ versus morphing angle $\gamma$ for each propeller failure case from Type 3 experiments.}\label{fig:rot_kin_en_circling}
\end{figure}

For all experiments, the translational error remains low, comparable to the healthy case.
Figure~\ref{fig:rot_kin_en_circling} presents the average $E_{\psi}^{\text{rot}}$ as a function of $\gamma$ for each failure case.
We observe that the rotational kinetic energy remains negligible, comparable to the nominal healthy case, for larger values of $|\gamma|$, indicating practical resilience.
However, these measures increase significantly when $|\gamma| < \pi/30$.
Consequently, we conclude that for this specific morphing prototype, all configurations satisfying $|\gamma| \geq \pi/30$ ($\approx 6^\circ$) consistently exhibit practical resilience.

\section{Opti-Hexa: Discussion and Conclusions}\label{sec:discussion}

This study was motivated by the historical trade-off between energy efficiency and practical resilience in hexarotor design. Prior to this work, the design landscape was largely polarized between the Star-shaped configuration ($\gamma = 0$), known for high efficiency but a lack of resilience, and the Y-shaped configuration ($\gamma = \pi/6$), known for resilience but poor efficiency.

Through the development and extensive testing of our open-source morphing platform, we have provided experimental proof of the existence of a viability region that bridges this gap.
By intersecting the results from our power consumption model (Section~\ref{sec:power_consumption}) and our resilience analysis (Section~\ref{sec:robust}), we identified a specific range of configurations for this particular platform, defined by the interval:
\begin{equation}
    (6^\circ \approx)\quad\frac{\pi}{30} \le |\gamma| \le \frac{\pi}{18}\quad(\approx10^\circ).
\end{equation}
We term any configuration within this interval as the \emph{Opti-Hexa}.
For values of $|\gamma| \le \pi/18$, the platform retains practical efficiency comparable to the standard Star-shape. Simultaneously, for values of $|\gamma| \ge \pi/30$, the platform exhibits practically resilient behavior, maintaining safe rotational kinetic energy and positioning accuracy after a propeller failure.
Consequently, the Opti-Hexa represents a physical realization of a platform that is, for the first time, experimentally verified to be both resilient and efficient.

This finding challenges the convention that resilience must come at the cost of significant efficiency loss. It suggests that the standard Star-shaped design is sub-optimal for safety-critical applications, as a minor geometric adjustment, for example shifting to $|\gamma| \ge \pi/30$ for our specific design, can confer resilience without degrading flight time.

We emphasize that this result is an experimental proof of existence for the specific hardware scale, mass, and propulsion system of our prototype. We do not claim that the specific angular bounds derived here ($\pi/30$ and $\pi/18$) act as universal constants for all multi-rotors. However, the methodology and the open-source morphing platform presented in this work may provide some of the tools for researchers and engineers to identify the optimal design---the Opti-Hexa---for their specific constraints, payloads, and scales.

\bibliographystyle{IEEEtran}
\bibliography{bibAlias,bibCustom,bibAF}

\end{document}